\documentclass[conference]{IEEEtran}
\IEEEoverridecommandlockouts
\usepackage{cite}

\usepackage{url}
\usepackage{amsmath,amssymb,amsfonts}
\usepackage{algorithmic}
\usepackage{graphicx}
\usepackage{textcomp}
\usepackage{xcolor}

\usepackage{microtype}

\usepackage{booktabs}
\usepackage{tabularx}

\def\BibTeX{{\rm B\kern-.05em{\sc i\kern-.025em b}\kern-.08em
    T\kern-.1667em\lower.7ex\hbox{E}\kern-.125emX}}

\begin{document}

\title{Pseudo-Labeling and Confirmation Bias in Deep Semi-Supervised Learning}

\author{%
Eric Arazo, Diego Ortego, Paul Albert, Noel E. O'Connor, Kevin McGuinness \\
  Insight Centre for Data Analytics, Dublin City University (DCU) \\
  \texttt{\{eric.arazo, diego.ortego\}}{@insight-centre.org} 
}

\maketitle

\begin{abstract}
Semi-supervised learning, i.e. jointly learning from labeled and unlabeled samples, is an active research topic due to its key role on relaxing human supervision. In the context of image classification, recent advances to learn from unlabeled samples are mainly focused on consistency regularization methods that encourage invariant predictions for different perturbations of unlabeled samples. We, conversely, propose to learn from unlabeled data by generating soft pseudo-labels using the network predictions. We show that a naive pseudo-labeling overfits to incorrect pseudo-labels due to the so-called confirmation bias and demonstrate that mixup augmentation and setting a minimum number of labeled samples per mini-batch are effective regularization techniques for reducing it. The proposed approach achieves state-of-the-art results in CIFAR-10/100, SVHN, and Mini-ImageNet despite being much simpler than other methods. These results demonstrate that pseudo-labeling alone can outperform consistency regularization methods, while the opposite was supposed in previous work.
Source code is available at \url{https://git.io/fjQsC}.
\end{abstract}

\section{Introduction\label{sec:Introduction}}

Convolutional neural networks (CNNs) have become the dominant approach in computer vision \cite{2017_ICCV_FocalLoss,2018_NIPS_FaceRecognition,2018_ECCV_Tracking,2018_ECCV_VideoClassif}. To best exploit them, vast amounts of labeled data are required. Obtaining such labels, however, is not trivial, and the research community is exploring alternatives to alleviate this~\cite{2017_arXiv_WebVision,2018_NIPS_RealisticEval,2019_TMAPI_SelfSupervised}.

Knowledge transfer via deep domain adaptation \cite{2018_Neurocomputing_DomainAdaptationSurvey} is a popular alternative that seeks to learn transferable representations from source to target domains by embedding domain adaptation in the learning pipeline. Other approaches focus exclusively on learning useful representations from scratch in a target domain when annotation constraints are relaxed \cite{2018_NIPS_RealisticEval,2019_ICML_LabelNoiseBMM,2018_ICLR_Rotation}. Semi-supervised learning (SSL) \cite{2018_NIPS_RealisticEval} focuses on scenarios with sparsely labeled data and extensive amounts of unlabeled data; learning with label noise \cite{2019_ICML_LabelNoiseBMM} seeks robust learning when labels are obtained automatically and may not represent the image content; and self-supervised learning \cite{2018_ICLR_Rotation} uses data supervision to learn from unlabeled data in a supervised manner. This paper focuses on SSL for image classification, a recently very active research area~\cite{2019_arXiv_CertaintyMT}. 

SSL is a transversal task for different domains including images~\cite{2018_NIPS_RealisticEval}, audio~\cite{2016_ICASSP_AudioVisualSSL}, time series~\cite{2018_KIS_SSLtimeSeries}, and text~\cite{2016_arXiv_SSLtext}. Recent approaches in image classification primarily focus on exploiting the consistency in the predictions for the same sample under different perturbations (consistency regularization) \cite{2016_NIPS_SSLperturbations,2019_arXiv_CertaintyMT},
while other approaches directly generate labels for the unlabeled data to guide the learning process (pseudo-labeling) \cite{2013_ICMLW_PseudoLab,2019_CVPR_LabelPropagation}. These two alternatives differ importantly in the mechanism they use to exploit unlabeled samples.
Consistency regularization and pseudo-labeling approaches apply different strategies such as a warm-up phase using labeled data \cite{2017_NIPS_MeanTeachers,2019_CVPR_LabelPropagation}, uncertainty weighting \cite{2018_ECCV_SSLtransductive,2019_arXiv_CertaintyMT}, adversarial attacks \cite{2018_TPAMI_VAT,2018_ECCV_DeepCoTraining}, or graph-consistency \cite{2018_CVPR_SmoothNeighGraphs,2019_CVPR_LabelPropagation}. These strategies deal with confirmation bias \cite{2017_NIPS_MeanTeachers, 2019_arXiv_CertaintyMT}, also known as noise accumulation \cite{2016_ICASSP_AudioVisualSSL}. This bias stems from using incorrect predictions on unlabeled data for training in subsequent epochs and, thereby increasing confidence in incorrect predictions and producing a model that will tend to resist new changes.


This paper explores pseudo-labeling for semi-supervised deep learning from the network predictions and shows that, contrary to previous attempts on pseudo-labeling \cite{2019_CVPR_LabelPropagation, 2018_NIPS_RealisticEval, 2018_ECCV_SSLtransductive}, simple modifications to prevent confirmation bias lead to state-of-the-art performance without adding consistency regularization strategies. As commonly done in the related literature \cite{2017_NIPS_MeanTeachers, 2018_TPAMI_VAT, 2019_CVPR_LabelPropagation, 2019_NIPS_MixMatch}, we focus the study on class-balanced scenarios. We adapt the approach proposed by Tanaka et al. \cite{2018_CVPR_JointOpt} in the context of label noise and apply it exclusively on unlabeled samples. Experiments show that this naive pseudo-labeling is limited by confirmation bias as prediction errors are fit by the network. To deal with this issue, we propose to use mixup augmentation \cite{2018_ICLR_mixup} as an effective regularization that helps calibrate deep neural networks~\cite{2019_arXiv_MIXUPCALIBRATE} and, therefore, alleviates confirmation bias. We find that mixup alone does not guarantee robustness against confirmation bias when reducing the amount of labeled samples or using certain network architectures (see Subsection \ref{subsec:GeneralizationArch}), and show that, when properly introduced, dropout regularization \cite{2014_JMLR_Droput} and data augmentation mitigates this issue. Our purely pseudo-labeling approach achieves state-of-the-art results (see Subsection \ref{subsec:SoTAcomparison}) without requiring multiple networks \cite{2017_NIPS_MeanTeachers,2018_ECCV_DeepCoTraining,2019_arXiv_CertaintyMT,2019_IJCAI_ICT}, nor does it require over a thousand epochs of training to achieve peak performance in every dataset \cite{2019_ICLR_sslSWA,2019_NIPS_MixMatch}, nor needs many (ten) forward passes for each sample \cite{2019_arXiv_CertaintyMT}. Compared to other pseudo-labeling approaches, the proposed approach is simpler in that it does not require graph construction and diffusion \cite{2019_CVPR_LabelPropagation} or combination with consistency regularization methods \cite{2018_ECCV_SSLtransductive}, but still achieves state-of-the-art results.

\section{Related work}

This section reviews closely related SSL methods, i.e. those using deep learning with mini-batch optimization over large image collections. Previous work on deep SSL differ in whether they use consistency regularization or pseudo-labeling to learn from the unlabeled set \cite{2019_CVPR_LabelPropagation},
while they all share the use of a cross-entropy loss (or similar) on labeled data.

\paragraph{\textbf{Consistency regularization}}

Imposes that the same sample under different perturbations must produce the same output. This idea was used in~\cite{2016_NIPS_SSLperturbations} where they apply randomized data augmentation, dropout, and random max-pooling while forcing softmax predictions to be similar. A similar idea is applied in~\cite{2017_ICLR_TemporalEnsemb}, which also extends the perturbation to different epochs, i.e. the current prediction for a sample has to be similar to an ensemble of predictions of the same sample in the past. Here the different perturbations come from networks at different states, dropout, and data augmentation. In~\cite{2017_NIPS_MeanTeachers}, the temporal ensembling method is interpreted as a teacher-student problem where the network is both a teacher that produces targets for the unlabeled data as a temporal ensemble, and a student that learns the generated targets by imposing the consistency regularization.~\cite{2017_NIPS_MeanTeachers}~naturally re-defines the problem to deal with confirmation bias by separating the teacher and the student. The teacher is defined as a different network with similar architecture whose parameters are updated as an exponential moving average of the student network weights. This method is extended in~\cite{2019_arXiv_CertaintyMT}, where they apply an uncertainty weight over the unlabeled samples to learn from the unlabeled samples with low uncertainty (i.e. entropy of the predictions for each sample under random perturbations). Additionally, Miyato et al.~\cite{2018_TPAMI_VAT} use virtual adversarial training to carefully introduce perturbations to data samples as adversarial noise and later impose consistency regularization on the predictions. More recently, Luo et al.~\cite{2018_CVPR_SmoothNeighGraphs} propose to use a contrastive loss on the predictions as a regularization that forces predictions to be similar (different) when they are from the same (different) class. This method extends the consistency regularization previously considered only in-between the same data samples to in-between different samples. Their method can naturally be combined with~\cite{2017_NIPS_MeanTeachers} or~\cite{2018_TPAMI_VAT} to boost their performance. Similarly, Verma et al.~\cite{2019_IJCAI_ICT} propose interpolation consistency training, a method inspired by \cite{2018_ICLR_mixup} that encourage predictions at interpolated unlabeled samples to be consistent with the interpolated predictions of individual samples. Also, authors in \cite{2019_NIPS_MixMatch} apply consistency regularization by guessing low-entropy labels, generating data-augmented unlabeled examples and mixing labeled and unlabeled examples using mixup \cite{2018_ICLR_mixup}. Both \cite{2019_IJCAI_ICT} and \cite{2019_NIPS_MixMatch} adopt \cite{2017_NIPS_MeanTeachers} to estimate the targets used in the consistency regularization.

Co-training~\cite{2018_ECCV_DeepCoTraining} uses two (or more) networks trained simultaneously to agree on their predictions (consistency regularization) and disagree on their errors. Errors are defined as different predictions when exposed to adversarial attacks, thus forcing different networks to learn complementary representations for the same samples. Recently, Chen et al.~\cite{2018_ECCV_SSLmemory} measure the consistency between the current prediction and an additional prediction for the same sample given by an external memory module that keeps track of previous representations. They additionally introduce an uncertainty weighting of the consistency term to reduce the contribution of uncertain predictions. Consistency regularization methods such as~\cite{2017_ICLR_TemporalEnsemb, 2017_NIPS_MeanTeachers, 2018_TPAMI_VAT} have all been shown to benefit from stochastic weight averaging method~\cite{2019_ICLR_sslSWA}, that averages network parameters at different training epochs to move the SGD solution on borders of flat loss regions to their center and improve generalization.

\paragraph{\textbf{Pseudo-labeling}}

Seeks the generation of labels or pseudo-labels for unlabeled samples to guide the learning process. An early attempt at pseudo-labeling proposed in \cite{2013_ICMLW_PseudoLab} uses the network predictions as labels. However, they constrain the pseudo-labeling to a fine-tuning stage, i.e. there is a pre-training or warm-up to initialize the network. A recent pseudo-labeling approach proposed in \cite{2018_ECCV_SSLtransductive} uses the network class prediction as hard labels for the unlabeled samples. They also introduce an uncertainty weight for each sample loss, it being higher for samples that have distant $k$-nearest neighbors in the feature space. They further include a loss term to encourage intra-class compactness and inter-class separation, and a consistency term between samples with different perturbations. Improved results are reported in combination with \cite{2017_NIPS_MeanTeachers}. Finally, a recently published work \cite{2019_CVPR_LabelPropagation} implements pseudo-labeling through graph-based label propagation. The method alternates between two steps: training from labeled and pseudo-labeled data and using the representations of the network  to build a nearest neighbor graph where label propagation is applied to refine hard pseudo-labels. They further add an uncertainty score for every sample (softmax prediction entropy based) and class (class population based) to deal, respectively, with the unequal confidence in network predictions and class-imbalance.

\section{Pseudo-labeling}

We formulate SSL as learning a model $h_{\theta}(x)$ from a set of $N$ training samples $\mathcal{D}$. These samples are split into the unlabeled set $\mathcal{D}_{u}=\left\{ x_{i}\right\} _{i=1}^{N_{u}}$ and the labeled set $\mathcal{D}_{l}=\left\{ \left(x_{i},y_{i}\right)\right\} _{i=1}^{N_{l}}$, being $y_{i}\in\left\{ 0,1\right\} ^{C}$ the one-hot encoding label for $C$ classes corresponding to $x_{i}$ and $N=N_{l}+N_{u}$. In our case, $h_{\theta}$ is a CNN and $\theta$ represents the model parameters (weights and biases). As we seek to perform pseudo-labeling, we assume that a pseudo-label $\tilde{y}$ is available for the $N_{u}$ unlabeled samples. We can then reformulate SSL as training using $\tilde{\mathcal{D}}=\left\{ \left(x_{i},\tilde{y}_{i}\right)\right\} _{i=1}^{N}$, being $\tilde{y}=y$ for the $N_{l}$ labeled samples.

The CNN parameters $\theta$ can be optimized using categorical cross-entropy: 
\begin{equation}
\ell^{*}(\theta)=-\sum_{i=1}^{N}\tilde{y}_{i}^{T}\log\left(h_{\theta}(x_{i})\right),\label{eq:Cross-ent}
\end{equation}
where $h_{\theta}(x)$ are the softmax probabilities produced by the model and $\log(\cdot)$ is applied element-wise. A key decision is how to generate the pseudo-labels $\tilde{y}$ for the $N_{u}$ unlabeled samples. Previous approaches have used hard pseudo-labels (i.e. one-hot vectors) directly using the network output class \cite{2013_ICMLW_PseudoLab,2018_ECCV_SSLtransductive} or the class estimated using label propagation on a nearest neighbor graph \cite{2019_CVPR_LabelPropagation}. We adopt the former approach, but use soft pseudo-labels, as we have seen this outperforms hard labels, confirming the observations noted in \cite{2018_CVPR_JointOpt} in the context of relabeling when learning with label noise. In particular, we store the softmax predictions $h_{\theta}(x_{i})$ of the network in every mini-batch of an epoch and use them to modify the soft pseudo-label $\tilde{y}$ for the $N_{u}$ unlabeled samples at the end of every epoch. We proceed as described from the second to the last training epoch, while in the first epoch we use the softmax predictions for the unlabeled samples from a model trained in a 10 epochs warm-up phase using the labeled data subset $\mathcal{D}_{u}$.

We use the two regularizations applied in \cite{2018_CVPR_JointOpt} to improve convergence. The first regularization deals with the difficulty of converging at early training stages when the network's predictions are mostly incorrect and the CNN tends to predict the same class to minimize the loss. Assignment of all samples to a single class is discouraged by adding: 
\begin{equation}
R_{A}=\sum_{c=1}^{C}p_{c}\log\left(\frac{p_{c}}{\overline{h}_{c}}\right),\label{eq:Regularization}
\end{equation}
where $p_{c}$ is the prior probability distribution for class $c$ and $\overline{h}_{c}$ denotes the mean softmax probability of the model for class $c$ across all samples in the dataset. As in \cite{2018_CVPR_JointOpt}, we assume a uniform distribution $p_{c}=1/C$ for the prior probabilities ($R_{A}$ stands for all classes regularization) and approximate $\overline{h}_{c}$ using mini-batches. The second regularization is needed to concentrate the probability distribution of each soft pseudo-label on a single class, thus avoiding the local optima in which the network might get stuck due to a weak guidance: 
\begin{equation}
R_{H}=-\frac{1}{N}\sum_{i=1}^{N}\sum_{c=1}^{C}h_{\theta}^{c}(x_{i})\log\left(h_{\theta}^{c}(x_{i})\right),\label{eq:Regularization-2}
\end{equation}
where $h_{\theta}^{c}(x_{i})$ denotes the $c$ class value of the softmax output $h_{\theta}(x_{i})$ and again using mini-batches (i.e. $N$ is replaced by the mini-batch size) to approximate this term. This second regularization is the average per-sample entropy ($R_{H}$ stands for entropy regularization), a well-known regularization in SSL \cite{2004_NIPS_EntropySSL}. Finally, the total semi-supervised loss is: 
\begin{equation}
\ell=\ell^{*}+\lambda_{A}R_{A}+\lambda_{H}R_{H},\label{eq:FullLoss}
\end{equation}
where $\lambda_{A}$ and $\lambda_{H}$ control the contribution of each regularization term (see Subsection~\ref{subsec:ExtendedHyperp} for a study of these hyperparameters). We stress that this pseudo-labeling approach adapted from \cite{2018_CVPR_JointOpt} is far from the state-of-the-art for SSL (see Subsection \ref{subsec:ConfBiasExp}), and are the mechanisms proposed in Subsection \ref{subsec:Confirmation-bias} which make pseudo-labeling a suitable alternative.

\subsection{Confirmation bias\label{subsec:Confirmation-bias}}
\begin{figure*}
\centering{}%
\begin{tabular}{ccc}
\includegraphics[width=0.25\textwidth,height=0.1\textheight]{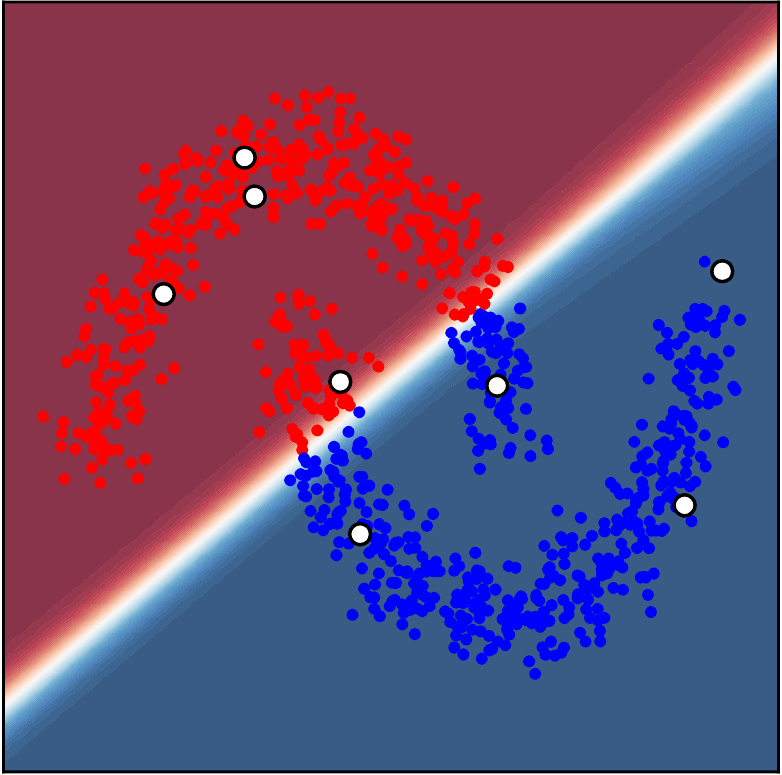} & \includegraphics[width=0.25\textwidth,height=0.1\textheight]{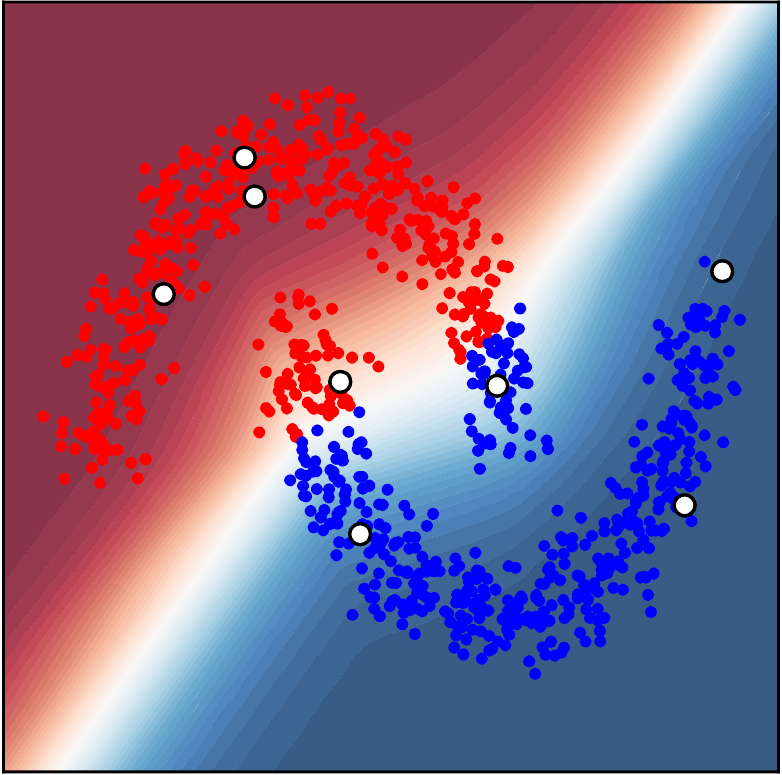} & \includegraphics[width=0.25\textwidth,height=0.1\textheight]{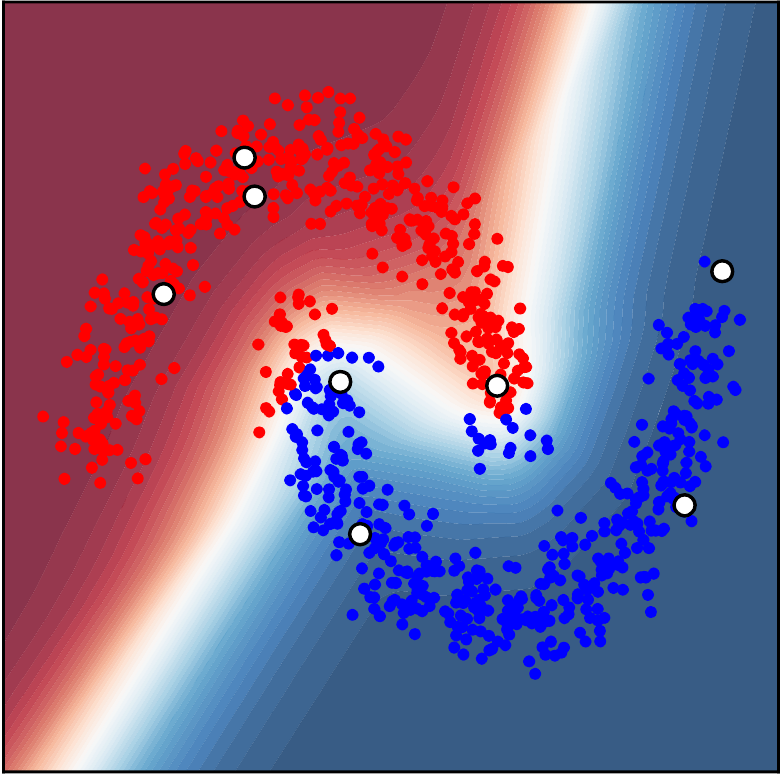}\tabularnewline
\end{tabular}\caption{\label{figTwoMoons}Pseudo-labeling in the ``two moons'' data (4 labels/class) for 1000 samples. From left to right: no mixup, mixup, and mixup with a minimum number of labeled samples per mini-batch. We use an NN classifier with one hidden layer with 50 hidden units as in \cite{2018_TPAMI_VAT}. Best viewed in color.}
\end{figure*}
Network predictions are, of course, sometimes incorrect. This situation is reinforced when incorrect predictions are used as labels for unlabeled samples, as it is the case in pseudo-labeling. Overfitting to incorrect pseudo-labels predicted by the network is known as confirmation bias. It is natural to think that reducing the confidence of the network on its predictions might alleviate this problem and improve generalization. Recently, mixup data augmentation \cite{2018_ICLR_mixup} introduced a strong regularization technique that combines data augmentation with label smoothing, which makes it potentially useful to deal with this bias. Mixup trains on convex combinations of sample pairs ($x_{p}$ and $x_{q}$) and corresponding labels ($y_{p}$ and $y_{q}$): 
\begin{equation}
x=\delta x_{p}+(1-\delta)x_{q},\label{eq:MixupData}
\end{equation}
\begin{equation}
y=\delta y_{p}+(1-\delta)y_{q},\label{eq:MixupLoss}
\end{equation}
where $\delta\in\left\{ 0,1\right\} $ is randomly sampled from a beta distribution $\mathcal{B}e\left(\alpha,\beta\right)$, with $\alpha=\beta$ (e.g. $\alpha=1$ uniformly selects $\delta$). This combination regularizes the network to favor linear behavior in-between training samples, reducing oscillations in regions far from them. Additionally, Eq. \ref{eq:MixupLoss} can be re-interpreted in the loss as $\ell^{*}=\delta\ell_{p}^{*}+(1-\delta)\ell_{q}^{*}$, thus re-defining the loss $\ell^{*}$ used in Eq. \ref{eq:FullLoss} as: 
\begin{equation}
\ell^{*}=-\sum_{i=1}^{N}\delta\left[\tilde{y}_{i,p}^{T}\log\left(h_{\theta}(x_{i})\right)\right]+\left(1-\delta\right)\left[\tilde{y}_{i,q}^{T}\log\left(h_{\theta}(x_{i})\right)\right].
\end{equation}
As shown in \cite{2019_arXiv_MIXUPCALIBRATE}, overconfidence in deep neural networks is a consequence of training on hard labels and it is the label smoothing effect from randomly combining $y_{p}$ and $y_{q}$ during mixup training that reduces prediction confidence and improves model calibration. In the semi-supervised context with pseudo-labeling, using soft-labels and mixup reduces overfitting to model predictions, which is especially important for unlabeled samples whose predictions are used as soft-labels. Note that training with mixup generates softmax outputs $h_{\theta}(x)$ for mixed inputs $x$, thus requiring a second forward pass with the original images to compute unmixed predictions. 

Mixup data augmentation alone may be insufficient to deal with confirmation bias when few labeled examples are provided. For example, when training with 500 labeled samples in CIFAR-10 and mini-batch size of 100, just 1 clean sample per batch is seen, which is especially problematic at early stages of training where little correct guidance is provided. Oversampling the labelled examples by setting a minimum number of labeled samples per mini-batch $k$ (as done in other works \cite{2017_NIPS_MeanTeachers, 2018_ECCV_SSLmemory, 2019_NIPS_MixMatch, 2019_CVPR_LabelPropagation}) provides a constant reinforcement with correct labels during training, reducing confirmation bias and helping to produce better pseudo-labels. 

The effect of this oversampling can be understood by splitting the total loss (Eq.~\ref{eq:Cross-ent}) into two terms, the first depending on the labeled examples and the second on the unlabelled:
\begin{equation}
\ell^{*}=N_{l}\overline{\ell}_{l}+N_{u}\overline{\ell}_{u},\label{eq:LabUnlabloss}
\end{equation}
where $N_{l}$ and $N_{u}$ are the number of labelled and unlabelled samples, and the  $\overline{\ell}_{l}=\frac{1}{N_{l}}\sum_{i=1}^{N_{l}}\ell_{l}^{(i)}$ is the average loss for labeled samples and similarly $\overline{\ell}_{u}$ for the unlabeled samples. The first term is a data loss on the labeled samples and the second can be interpreted as a regularization term that encourages the network to fit the pseudo-labels of the unlabeled samples. When few labeled samples are available, $N_{l}<<N_{u}$, the regularization term dominates the loss, i.e. fitting the pseudo-labels is weighted far higher than fitting the labelled samples. This can be overcome either by upweighting the the first term or by oversampling labeled samples. We use the latter strategy as it results in more frequent parameter updates to satisfy the first term, rather than larger magnitude updates.
%
%
%
Subsections \ref{subsec:ConfBiasExp} and \ref{subsec:GeneralizationArch} experimentally show that mixup, a minimum number of samples per mini-batch, and other techniques (dropout and data augmentation) reduce confirmation bias and make pseudo-labeling an effective alternative to consistency regularization.

\section{Experimental work}

\subsection{Datasets and training\label{subsec:Datasets-and-training}}
We use four image classification datasets, CIFAR-10/100~\cite{2009_CIFAR}, SVHN~\cite{2011_NeurIPS_SVHN} and Mini-ImageNet~\cite{2016_NIPS_MiniImageNet}, to validate our approach. Part of the training images are labeled and the remaining are unlabeled. Following \cite{2018_NIPS_RealisticEval}, we use a validation set of 5K samples for CIFAR-10/100 for studying hyperparameters in Subsections \ref{subsec:ConfBiasExp} and \ref{subsec:GeneralizationArch}. However, as done in \cite{2019_ICLR_sslSWA}, we add the 5K samples back to the training set for comparisons in Subsection \ref{subsec:SoTAcomparison}, where we report test results (model from the best epoch).

\paragraph{\textbf{CIFAR-10, CIFAR-100, and SVHN}}
These datasets contain 10, 100, and 10 classes respectivelly,  with 50K color images for training and 10K for testing in CIFAR-10/100 and 73257 images for training and 26032 for testing in SVHN. The three datasets have resolution 32\texttimes 32. We perform experiments with a number of labeled images $N_{l}=$ 0.25K, 0.5K, and 1K for SVHN and $N_{l}=$ 0.25K, 0.5K, 1K, and 4K (4K and 10K) for CIFAR-10 (CIFAR-100). We use the well-known ``13-CNN'' architecture \cite{2019_ICLR_sslSWA} for CIFAR-10/100 and SVHN. We also experiment with a Wide ResNet-28-2 (WR-28) \cite{2018_NIPS_RealisticEval} and a PreAct ResNet-18 (PR-18) \cite{2018_ICLR_mixup} in Subsection \ref{subsec:GeneralizationArch} to study the generalization to different architectures.

\paragraph{\textbf{Mini-ImageNet}}
We emulate the semi-supervised learning setup Mini-ImageNet~\cite{2016_NIPS_MiniImageNet}
(a subset of the well-known ImageNet \cite{2009_CVPR_ImageNet} dataset) used in \cite{2019_CVPR_LabelPropagation}. Train and test sets of 100 classes and 600 color images per class with resolution 84 \texttimes{} 84 are selected from ImageNet, as in \cite{2017_ICLR_MiniImNetsplits}. 500 (100) images per-class are kept for train (test) splits. The train and test sets therefore contain 50k and 10k images. As with CIFAR-100, we experiment with a number of labeled images $N_{l}=$ 4K and 10K. Following \cite{2019_CVPR_LabelPropagation}, we use a ResNet-18 (RN-18) architecture~\cite{2016_CVPR_ResNet}.

\begin{figure*}[t]
\centering{}%
\begin{tabular}{cc}
\includegraphics[width=0.45\textwidth]{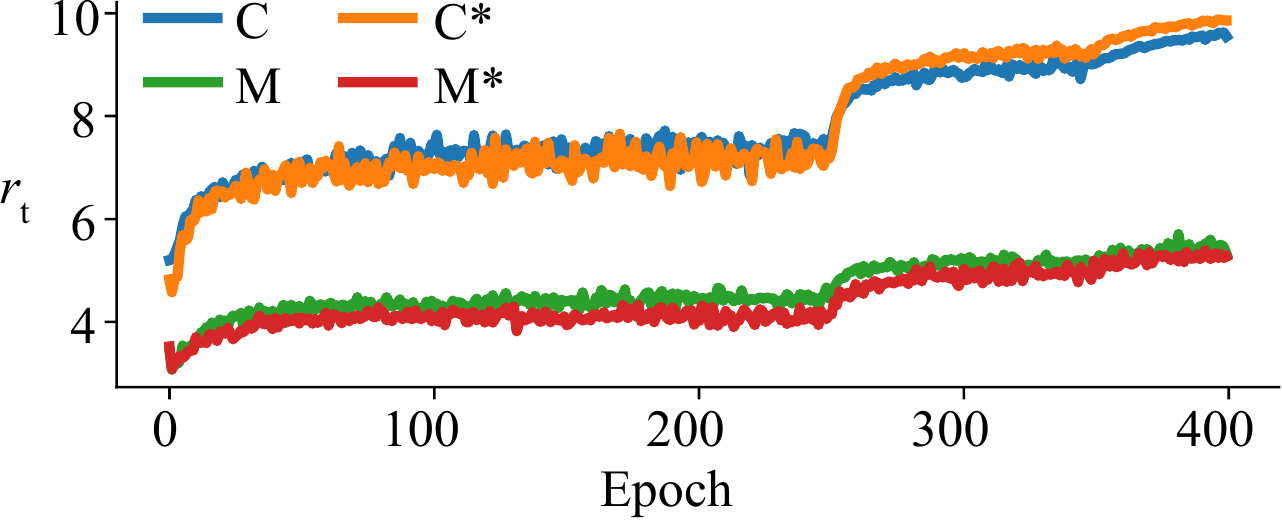} & \includegraphics[width=0.45\textwidth]{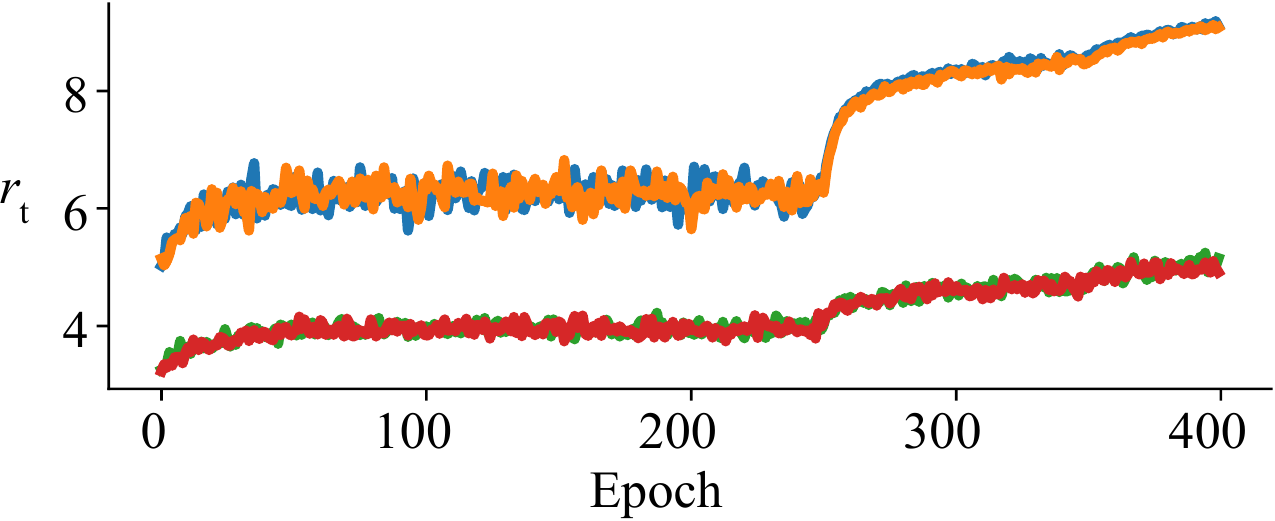}\tabularnewline
\end{tabular}\caption{\label{fig:ConfirmationBiasFigure}Example of certainty of incorrect predictions $r_{t}$ during training when using 500 (left) and 4000 (right) labeled images in CIFAR-10. Moving from cross-entropy (C) to mixup (M) reduces $r_{t}$, whereas adding a minimum number of samples per mini-batch ({*}) also helps in 500 labels, where M{*} (with slightly lower $r_{t}$ than M) is the only configuration that converges, as shown in Table \ref{tab:Mixup-and-mini-batch} (top). Best viewed in color.}
\end{figure*}
\paragraph{\textbf{Hyperparameters}}
We use the typical configuration for CIFAR-10/100 and SVHN \cite{2017_ICLR_TemporalEnsemb}, and the same for Mini-ImageNet. Image normalization using dataset mean and standard deviation and subsequent data augmentation \cite{2017_ICLR_TemporalEnsemb} by random horizontal flips and 2 (6) pixel translations for CIFAR and SVHN (Mini-ImageNet). Additionally, color jitter is applied as in \cite{2019_ICCV_SurprisingDA} in Subsections \ref{subsec:GeneralizationArch} and \ref{subsec:SoTAcomparison} for higher robustness against confirmation bias. We train using SGD with momentum of 0.9, weight decay of $10^{-4}$, and batch size of 100. Training always starts with a high learning rate (0.1 in CIFAR and SVHN, and 0.2 in Mini-ImageNet), dividing it by ten twice during training. We train for CIFAR and Mini-ImageNet 400 epochs (reducing learning rate in epochs 250 and 350) and use 10 epoch warm-up with labeled data, while for SVHN we train 150 epochs (reducing learning rate in epochs 50 and 100) and use a longer warm-up of 150 epochs to start the pseudo-labeling with good predictions and leading to reliable convergence (experiments in CIFAR-10 with longer warm-up provided results in the same error range already reported). We do not attempt careful tuning of the regularization weights $\lambda_{A}$ and $\lambda_{H}$ and just set them to 0.8 and 0.4 as done in \cite{2018_CVPR_JointOpt} (see Subsection~\ref{subsec:ExtendedHyperp} for an ablation study of these parameters). When using dropout, it is introduced between consecutive convolutional layers of ResNet blocks in WR-28, PR-18, and RN-18, while for 13-CNN we introduce it as in \cite{2017_ICLR_TemporalEnsemb}. Following \cite{2019_ICLR_sslSWA}\footnote{\url{https://github.com/benathi/fastswa-semi-sup}}, we use weight normalization \cite{2016_NIPS_WeightNorm} in all networks.
\subsection{Effect of mixup on confirmation bias\label{subsec:ConfBiasExp}}
This section demonstrates that carefully regularized pseudo-labeling is a suitable alternative for SSL. Figure~\ref{figTwoMoons} illustrates  our approach on the ``two moons'' toy data. Figure~\ref{figTwoMoons} (left) shows the limitations of a naive pseudo-labeling adapted from \cite{2018_CVPR_JointOpt}, which fails to adapt to the structure in the unlabelled examples and results in a linear decision boundary. Figure~\ref{figTwoMoons} (middle) shows the effect
of mixup, which alleviates confirmation bias to better model the structure and gives a smoother boundary. Figure~\ref{figTwoMoons} (right) shows that combining mixup  with a minimum number of labeled samples $k$ per mini-batch improves the semi-supervised decision boundary.

\begin{table}
\centering{}\caption{Confirmation bias alleviation using mixup and a minimum number of $k$ labeled samples per mini-batch. Top: Validation error for naive pseudo-labeling without mixup (C), mixup (M), and alternatives with minimum $k$. Bottom: Study of the effect of $k$ on the validation error. \label{tab:Mixup-and-mini-batch}}
\begin{tabularx}{\columnwidth}{Xccc}
\toprule
\multicolumn{1}{l}{} & \multicolumn{2}{c}{CIFAR-10} & \multicolumn{1}{c}{CIFAR-100}\tabularnewline
\midrule 
Labeled images & 500 & 4000 & 4000\tabularnewline
\midrule 
C & 52.44 & 11.40 & 48.54\tabularnewline
C{*} $\left(k=16\right)$ & 35.08 & 10.90 & 46.60\tabularnewline
M & 32.10 & 7.16 & 41.80\tabularnewline
M {*}$\left(k=16\right)$ & \textbf{13.68} & \textbf{6.90} & \textbf{38.78}\tabularnewline
\bottomrule 
&&&\tabularnewline  
\toprule
\multicolumn{1}{l}{} & \multicolumn{2}{c}{CIFAR-10} & \multicolumn{1}{c}{CIFAR-100}\tabularnewline
\midrule 
Labeled images & 500 & 4000 & 4000\tabularnewline
\midrule 
$k=8$ & \textbf{13.14} & 7.18 & 42.32\tabularnewline
$k=16$ & 13.68 & \textbf{6.90} & \textbf{38.78}\tabularnewline
$k=32$ & 14.58 & 7.06 & 39.62\tabularnewline
$k=64$ & 19.40 & 8.20 & 46.28\tabularnewline
\bottomrule 
\end{tabularx}\tabularnewline
\end{table}

Naive pseudo-labeling leads to overfitting the network predictions and high training accuracy in CIFAR-10/100. Table~\ref{tab:Mixup-and-mini-batch} (top) reports mixup effect in terms of validation error. Naive pseudo-labeling leads to an error of 11.40/48.54 for CIFAR-10/100 when training with cross-entropy (C) loss for 4000 labels. This error can be greatly reduced when using mixup (M) to 7.16/41.80. However, when further reducing the number of labels to 500 in CIFAR-10, M is insufficient to ensure low-error (32.10). We propose to set a minimum number of samples $k$ per mini-batch to tackle the problem. Table~\ref{tab:Mixup-and-mini-batch} (bottom) studies this parameter $k$ when combined with mixup, showing that 16 samples per mini-batch works well for both CIFAR-10 and CIFAR-100, dramatically reducing  error in all cases (e.g. in CIFAR-10 for 500 labels error is reduced from 32.10 to 13.68).
%
%
Confirmation bias causes a dramatic increase in the certainty of incorrect predictions during training. To demonstrate this behavior we compute the average cross-entropy of the softmax output with a uniform distribution $\mathcal{U}$, across the classes in every epoch $t$ for all incorrectly predicted samples $\left\{ x_{m_{t}}\right\} _{m_{t}=1}^{M_{t}}$ as: $r_{t}=-\frac{1}{M_{t}}\sum_{m_{t}=1}^{M_{t}}\mathcal{U}^{T}\log\left(h_{\theta}(x_{m_{t}})\right)$, where ${M_{t}}$ is the number of incorrectly predicted samples. Figure~\ref{fig:ConfirmationBiasFigure} shows that mixup and minimum $k$ are effective regularizers for reducing $r_{t}$, i.e. confirmation bias is reduced. We also experimented with using label noise regularizations \cite{2016_CVPR_NoiseLayer}, but setting a minimum $k$ proved more effective.
\subsection{Extended hyperparameters study\label{subsec:ExtendedHyperp}}

This subsection studies the effect of $\alpha$, $\lambda_A$, and $\lambda_H$ hyperparameters of our pseudo-labeling approach. Table~\ref{tab:alphaMixup_lambda} reports the validation error in CIFAR-10 using 500 and 4000 labels for, respectively, $\alpha$ and $\lambda_A$ and $\lambda_H$ . Note that we keep the same configuration used in Subsection~\ref{subsec:ConfBiasExp} with $k=16$, i.e. no dropout or additional data augmentation is used.
Table~\ref{tab:alphaMixup_lambda} results suggest that $\alpha = 4$ and $\alpha = 8$ values might further improve the reported results using $\alpha = 1$. However, we experimented on CIFAR-10 with 500 labels using the final configuration (adding dropout and additional data augmentation) and observed marginal differences (8.54 with $\alpha = 4$, which is within the error range of the 8.80 $\pm$ 0.45 obtained with $\alpha = 1$) shown in Table~\ref{tab:ValidationErrorCIFAR10-100}, thus suggesting that stronger mixup regularization might not be additive to dropout and extra data augmentation in our case. Table~\ref{tab:alphaMixup_lambda} shows that our configuration ($\lambda_A=0.8$ and $\lambda_H=0.4$) adopted from~\cite{2018_CVPR_JointOpt} is very close to the best performance in this experiment where marginal improvements are achieved. More careful hyperparameter tuning might slightly improve the results here, but the default configuration is already good and generalizes well across datasets.

\begin{table*}[t]
\begin{centering}
\caption{\label{tab:alphaMixup_lambda}Validation error for different values of the $\alpha$ parameter from Mixup, $\lambda_A$, and $\lambda_H$. Bold indicates lowest error. Underlined values indicate the results of the configuration used.}
\par\end{centering}
\begin{centering}
\begin{tabular}{ccccccccc}
\toprule
\multicolumn{1}{c}{Labeled images:} & \multicolumn{4}{c|}{500} & \multicolumn{4}{c}{4000}\tabularnewline
\midrule 
$\alpha$ & 0.1 & 1 & 4 & \multicolumn{1}{c|}{8} & 0.1 & 1 & 4 & 8\tabularnewline
\midrule 
 & 23.18 & \underline{13.68} & \textbf{10.60} & 11.04 & 8.58 & \underline{6.90} & \textbf{6.56} & 6.68\tabularnewline
\midrule
\midrule
$\lambda_A / \lambda_H$ & 0.1 & 0.4 & 0.8 & \multicolumn{1}{c|}{2} & 0.1 & 0.4 & 0.8 & 2\tabularnewline
\midrule 
0.1 & 22.94 & 29.64 & 60.76 & 83.96 & 7.22 & \textbf{6.88} & 7.74 & 33.98\tabularnewline
0.4 & 20.92 & \textbf{12.88} & 17.62 & 38.40 & 7.18 & 6.96 & 7.18 & 8.82\tabularnewline
0.8 & 23.50 & \underline{13.68} & 14.72 & 25.92 & 7.24 & \underline{6.90} & 7.18 & 8.78\tabularnewline
2 & 31.30 & 14.80 & 14.62 & 23.40 & 8.16 & 7.28 & 7.40 & 8.64\tabularnewline
\bottomrule 
\end{tabular}
\par\end{centering}
\end{table*}

\subsection{Generalization to different architectures\label{subsec:GeneralizationArch}}

There are examples in the recent literature~\cite{2019_CVPR_RevisitingSelfSup} where moving from one architecture to another changes which methods appear to have a higher potential. Kolesnikov et al.~\cite{2019_CVPR_RevisitingSelfSup} show that skip-connections in ResNet architectures play a key role on the quality of learned representations, while most approaches in previous literature were systematically evaluated using AlexNet \cite{2012_NIPS_AlexNet}. Ulyanov et al.~\cite{2018_CVPR_DeepImPrior} showed that different architectures lead different and useful image priors, highlighting the importance of exploring different networks. We, therefore, test our method with two more architectures: a Wide ResNet-28-2 (WR-28) \cite{2016_BMVC_WideResNet} typically used in SSL \cite{2018_NIPS_RealisticEval} (1.5M parameters) and a PreAct ResNet-18 (PR-18) \cite{2016_ECCV_PreActResNet} used in the context of label noise \cite{2018_ICLR_mixup} (11M parameters). Table~\ref{tab:GeneralizationArch} presents the results for the 13-CNN (AlexNet-type) and these network architectures (ResNet-type). Our pseudo-labeling with mixup and $k=16$ (M{*}) works well for 4000 and 500 labels across architectures, except for 500 labels for WR-28 where there is large error increase (29.50). This is due to a stronger confirmation bias in which labeled samples are not properly learned, while incorrect pseudo-labels are fit. Interestingly, PR-18 (11M) is more robust to confirmation bias than WR-28 (1.5M), while the 13-layer network (3M) has fewer parameters than PR-18 and achieves better performance. This suggests that the network architecture plays an important role, being a relevant prior for SSL with few labels.

\begin{table}[t]
\centering{}\caption{\label{tab:GeneralizationArch}Validation error across architectures is stabilized using dropout $p$ and data augmentation (A).}
\begin{tabularx}{\columnwidth}{Xcc}
\toprule
Labeled images & 500 & 4000\tabularnewline
\midrule 
\multicolumn{3}{l}{13-layer}\tabularnewline
\midrule 
M{*} & 13.68 & 6.90 \tabularnewline
M{*} $(p=0.1)$ & 12.62 & 6.58 \tabularnewline
M{*} $(p=0.3)$ & 11.94 & 6.66 \tabularnewline
M{*} $(p=0.1, A)$ &\textbf{9.16} & \textbf{6.22}\tabularnewline
\midrule 
\multicolumn{3}{l}{WR-28}\tabularnewline
\midrule 
M{*} & 29.50 & \textbf{6.40} \tabularnewline
M{*} $(p=0.1)$ & 14.14 & 7.06 \tabularnewline
M{*} $(p=0.3)$ & 30.56 & 11.44 \tabularnewline
M{*} $(p=0.1, A)$ & \textbf{10.94} & 6.74 \tabularnewline
\midrule 
\multicolumn{3}{l}{PR-18}\tabularnewline
\midrule 
M{*} & \textbf{13.90} & \textbf{5.94} \tabularnewline
M{*} $(p=0.1)$ & 14.78 & 5.90 \tabularnewline
M{*} $(p=0.3)$ & 14.78 & 6.62 \tabularnewline
M{*} $(p=0.1, A)$ & 14.96 & 6.32 \tabularnewline
\bottomrule 
\end{tabularx}\tabularnewline
\end{table}

We found that dropout \cite{2014_JMLR_Droput} and data augmentation help to achieve good performance across all architectures. Table~\ref{tab:GeneralizationArch} shows that dropout \emph{$p=0.1,0.3$} helps in achieving better convergence in CIFAR-10, whereas adding color jitter as additional data augmentation (details in Subsection \ref{subsec:Datasets-and-training}) further contributes to error reduction. Note that the quality of pseudo-labels is key, so it is essential to disable dropout to prevent corruption when computing these in the second forward pass. We similarly disable data augmentation in the second forward pass, which consistently improves performance. This configuration is used for comparison with the state-of-the-art in Subsection \ref{subsec:SoTAcomparison}.

\subsection{Comparison with the state-of-the-art\label{subsec:SoTAcomparison}}
We compare our pseudo-labeling approach against related work that makes use of the 13-CNN \cite{2017_NIPS_MeanTeachers} in CIFAR-10/100: $\varPi$ model \cite{2017_ICLR_TemporalEnsemb}, TE \cite{2017_ICLR_TemporalEnsemb}, MT \cite{2017_NIPS_MeanTeachers}, $\varPi$ model-SN \cite{2018_CVPR_SmoothNeighGraphs}, MA-DNN \cite{2018_ECCV_SSLmemory}, Deep-Co \cite{2018_ECCV_DeepCoTraining}, TSSDL \cite{2018_ECCV_SSLtransductive}, LP \cite{2019_CVPR_LabelPropagation}, CCL \cite{2019_arXiv_CertaintyMT}, fast-SWA \cite{2019_ICLR_sslSWA} and ICT \cite{2019_IJCAI_ICT}. Tables~\ref{tab:ValidationErrorCIFAR10-100} and~\ref{tab:ValidationErrorSVHN} divide methods into those based on consistency regularization and pseudo-labeling. Note that we include pseudo-labeling approaches combined with consistency regularization ones (e.g. MT) in the consistency regularization set. The proposed approach clearly outperforms consistency regularization methods, as well as other purely pseudo-labeling approaches and their combination with consistency regularization methods in CIFAR-10/100. In SVHN our pseudo-labeling approach outperforms most state-of-the-art methods, especially when there are very few labels. These results demonstrate the generalization of the proposed approach compared to other methods that fail when decreasing the number of labels. Furthermore, Table \ref{tab:ValidationErrorLowCIFAR10-MiniIm} (left) demonstrates that the proposed approach successfully scales to higher resolution images, obtaining an over 10 point margin on the best related work in Mini-ImageNet. Note that all supervised baselines are reported using the same data augmentation and dropout as in the proposed pseudo-labeling.

\begin{table}[t]
\centering{}\caption{\label{tab:ValidationErrorSVHN}Test error in SVHN for the proposed approach using the 13-CNN network. ({*}) denotes that we have run
the algorithm. Bold indicates lowest error. We report average and
standard deviation of 3 runs with different labeled/unlabeled splits.}
\begin{tabularx}{\columnwidth}{Xccc}
\toprule 
Labeled images & 250 & 500 & 1000 \tabularnewline
\midrule 
Supervised (C){*} & 43.60$\pm$3.35 & 22.67$\pm$2.80 & 13.32$\pm$0.89\tabularnewline
Supervised (M){*} & 53.15$\pm$6.54 & 20.74$\pm$0.80 & 11.66$\pm$0.17\tabularnewline
\midrule 
\multicolumn{4}{c}{Consistency regularization methods}\tabularnewline
\midrule 
$\varPi$ model & 9.69 $\pm$ 0.92 & 6.83 $\pm$ 0.66 & 4.95 $\pm$ 0.26\tabularnewline
TE & - & 5.12 $\pm$ 0.13 & 4.42 $\pm$ 0.16\tabularnewline
MT & 4.35 $\pm$ 0.50 & 4.18 $\pm$ 0.27 & 3.95 $\pm$ 0.19\tabularnewline
$\varPi$ model-SN & 5.07 $\pm$ 0.25 & 4.52 $\pm$ 0.30 & 3.82 $\pm$ 0.25\tabularnewline
MA-DNN & - & - & 4.21 $\pm$ 0.12\tabularnewline
Deep-Co & - & - & 3.61 $\pm$ 0.15\tabularnewline
MT-TSSDL & 4.09 $\pm$ 0.42 & 3.90 $\pm$ 0.27 & \textbf{3.35 $\pm$ 0.27}\tabularnewline
ICT & 4.78 $\pm$ 0.68 & 4.23 $\pm$ 0.15 & 3.89 $\pm$ 0.04\tabularnewline
\midrule 
\multicolumn{4}{c}{Pseudo-labeling methods}\tabularnewline
\midrule 
TSSDL & 5.02 $\pm$ 0.26 & 4.32 $\pm$ 0.30 & 3.80 $\pm$ 0.27\tabularnewline
Ours{*} & \textbf{3.66 $\pm$ 0.12} & \textbf{3.64 $\pm$ 0.04} & 3.55 $\pm$ 0.08\tabularnewline
\bottomrule
\end{tabularx}
\end{table}
\begin{table*}[t]
\centering{}\caption{\label{tab:ValidationErrorCIFAR10-100}Test error in CIFAR-10/100 for the proposed approach using the 13-CNN network. ({*}) denotes that we have run the algorithm. Bold indicates lowest error. We report average and standard deviation of 3 runs with different labeled/unlabeled splits.}
\begin{tabular}{lccccc}
\toprule 
 & \multicolumn{3}{c}{CIFAR-10} & \multicolumn{2}{c}{CIFAR-100} \tabularnewline
\midrule 
Labeled images & 500 & 1000 & 4000 & 4000 & 10000 \tabularnewline
\midrule 
Supervised (C){*} & 43.64 $\pm$ 1.21  & 34.83 $\pm$ 1.15 & 19.26 $\pm$ 0.26 & 54.49 $\pm$ 0.53 & 41.14 $\pm$ 0.26 \tabularnewline
Supervised (M){*} & 37.60 $\pm$ 0.65  & 28.59 $\pm$ 1.21  & 15.94 $\pm$ 0.26  & 52.70 $\pm$ 0.28  & 39.42 $\pm$ 0.37  \tabularnewline
\midrule 
\multicolumn{6}{c}{Consistency regularization methods}\tabularnewline
\midrule 
$\varPi$ model & - & - & 12.36 $\pm$ 0.31 & - & 39.19 $\pm$ 0.36  \tabularnewline
TE & - & - & 12.16 $\pm$ 0.24 & - & 38.65 $\pm$ 0.51  \tabularnewline
MT & 27.45 $\pm$ 2.64 & 19.04 $\pm$ 0.51 & 11.41 $\pm$ 0.25 & 45.36 $\pm$ 0.49 & 36.08 $\pm$ 0.51 \tabularnewline
$\varPi$ model-SN & - & 21.23 $\pm$ 1.27 & 11.00 $\pm$ 0.13 & - & 37.97 $\pm$ 0.29  \tabularnewline
MA-DNN & - & - & 11.91 $\pm$ 0.22 & - & 34.51 $\pm$ 0.61  \tabularnewline
Deep-Co & - & - & 9.03 $\pm$ 0.18 & - & 38.77 $\pm$ 0.28  \tabularnewline
MT-TSSDL & - & 18.41 $\pm$ 0.92 & 9.30 $\pm$ 0.55 & - & -   \tabularnewline
MT-LP & 24.02 $\pm$ 2.44 & 16.93 $\pm$ 0.70 & 10.61 $\pm$ 0.28 & 43.73 $\pm$ 0.20 & 35.92 $\pm$ 0.47  \tabularnewline
MT-CCL & - & 16.99 $\pm$ 0.71 & 10.63 $\pm$ 0.22 & - & 34.81 $\pm$ 0.52  \tabularnewline
MT-fast-SWA & - & 15.58 $\pm$ 0.12 & 9.05 $\pm$ 0.21 & - & 34.10 $\pm$ 0.31  \tabularnewline
ICT & - & 15.48 $\pm$ 0.78 & 7.29 $\pm$ 0.02 & - & - \tabularnewline
\midrule 
\multicolumn{6}{c}{Pseudo-labeling methods}\tabularnewline
\midrule 
TSSDL & - & 21.13 $\pm$ 1.17 & 10.90 $\pm$ 0.23 & - & -  \tabularnewline
LP & 32.40 \textbf{$\pm$ }1.80 & 22.02 $\pm$ 0.88 & 12.69 $\pm$ 0.29 & 46.20 $\pm$ 0.76 & 38.43 $\pm$ 1.88  \tabularnewline
Ours{*} & \textbf{8.80 $\pm$ 0.45 } & \textbf{6.85 $\pm$ 0.15} & \textbf{5.97 $\pm$ 0.15} & \textbf{37.55 $\pm$ 1.09} & \textbf{32.15 $\pm$ 0.50} \tabularnewline
\bottomrule
\end{tabular}
\end{table*}
%
%
%
\begin{table*}[t]
\centering{}\caption{\label{tab:ValidationErrorLowCIFAR10-MiniIm}Test error in Mini-ImageNet (left) and CIFAR-10 with few labeled samples (right). ({*}) denotes
that we have run the algorithm. Bold indicates lowest error. We report
average and standard deviation of 3 runs with different labeled/unlabeled
splits.}
\begin{tabular}{lc}
\begin{tabular}{lcc}
\toprule 
Labeled images & 4000 & 10000\tabularnewline
\midrule 
Supervised (C){*}  & 75.69 $\pm$ 0.24 & 63.24 $\pm$ 0.33 \tabularnewline
Supervised (M){*} & 72.03 $\pm$ 0.21 & 59.96 $\pm$ 0.40\tabularnewline
\midrule 
\multicolumn{3}{c}{Consistency regularization methods}\tabularnewline
\midrule
MT & 72.51 $\pm$ 0.22 & 57.55 $\pm$ 1.11\tabularnewline
MT-LP & 72.78 $\pm$ 0.15 & 57.35 $\pm$ 1.66\tabularnewline
\midrule 
\multicolumn{3}{c}{Pseudo-labeling methods}\tabularnewline
\midrule
LP & 70.29 $\pm$ 0.81 & 57.58 $\pm$ 1.47\tabularnewline
Ours{*} & \textbf{56.49 $\pm$ 0.51} & \textbf{46.08 $\pm$ 0.11}\tabularnewline
\bottomrule
\end{tabular}
 & %
\begin{tabular}{lccc}
\toprule 
Labeled images & 250 & 500 & 4000\tabularnewline
\midrule
MM (WR-28) & \textbf{11.08 $\pm$ 0.87} & \textbf{9.65 $\pm$ 0.94} & \textbf{6.24 $\pm$ 0.06}\tabularnewline
ICT{*} (WR-28) & 52.19 $\pm$ 1.54 & 42.33 $\pm$ 0.08 & 7.26 $\pm$ 0.04\tabularnewline
Ours{*} (WR-28) & 24.81 $\pm$ 5.35 & 14.25 $\pm$ 0.86 & 6.28 $\pm$ 0.3\tabularnewline
\midrule
Ours{*} (13-CNN) & \textbf{9.37 $\pm$ 0.12} & \textbf{8.80 $\pm$ 0.45}  & 5.97 $\pm$ 0.15\tabularnewline
Ours{*} (PR-18) & 23.86 $\pm$ 4.82 & 12.16 $\pm$ 1.06 & \textbf{5.86 $\pm$ 0.17}\tabularnewline
\bottomrule
\end{tabular}\tabularnewline
\end{tabular}
\end{table*}

Table~\ref{tab:ValidationErrorLowCIFAR10-MiniIm} (right) compares our pseudo-labeling approach against recent consistency regularization approaches that use mixup.
We achieve better performance than ICT \cite{2019_IJCAI_ICT}, while being competitive with MM \cite{2019_NIPS_MixMatch} for 500 and 4000 labels using WR-28. Regarding PR-18, we converge to reasonable performance for 4000 and 500 labels, whereas for 250 we do not. Finally, the 13-CNN robustly converges even for 250 labels where we obtain 9.37 test error. Therefore, these results suggest that it is worth exploring the relationship between number of labels, dataset complexity and architecture type. As shown in Subsection \ref{subsec:GeneralizationArch}, dropout and additional data augmentation help with 500 labels/class across architectures, but are insufficient for 250 labels. Better data augmentation \cite{2019_ICML_PBAugm} or self-supervised pre-training \cite{2019_ICCV_SelfSup} might overcome this challenge. 
However,  it is already interesting that a straightforward modification of pseudo-labeling, designed to tackle confirmation bias, gives a competitive  semi-supervised learning approach, without any consistency regularization, and future work should take this into account.

\section{Conclusions}

This paper presented a semi-supervised learning approach for image classification based on pseudo-labeling. We proposed to directly use the network predictions as soft pseudo-labels for unlabeled data together with mixup augmentation, a minimum number of labeled samples per mini-batch, dropout and data augmentation to alleviate confirmation bias. This conceptually simple approach outperforms related work in four datasets, demonstrating that pseudo-labeling is a suitable alternative to the dominant approach in recent literature: consistency-regularization. The proposed approach is, to the best of our knowledge, both simpler and more accurate than most recent approaches. Future work should explore SSL in class-unbalanced and large-scale datasets and synergies of pseudo-labelling and consistency regularization.

\section*{Acknowledgment}
This publication has emanated from research conducted with the financial support of Science Foundation Ireland (SFI) under grant number SFI/15/SIRG/3283 and SFI/12/RC/2289\_P2.

\bibliographystyle{IEEEtran}
\bibliography{bibliography_IJCNN2020.bib}


\end{document}